\documentclass{article} 
\usepackage{iclr2016_conference,times}
\usepackage{amsmath,amssymb} 
\usepackage{graphicx}
\usepackage{hyperref}
\usepackage{url}
\usepackage{subcaption}
\usepackage{soul}
\usepackage{gensymb}

\newcommand{\A}{A}
\newcommand{\B}{B}

\newcommand{\C}{\mathcal{C}}
\newcommand{\Q}{\mathcal{Q}}

\newcommand{\future}[1]{} 
\newcommand{\comment}[1]{}

\newcommand{\fig}[1]{Figure~\ref{fig:#1}}

\newcommand{\later}[1]{}
\newcommand{\longer}[1]{}	

\def\BE{\begin{equation}}
\def\EE{\end{equation}}

\def\BEA{\begin{eqnarray}}
\def\EEA{\end{eqnarray}}
\def\BEAS{\begin{eqnarray*}}
\def\EEAS{\end{eqnarray*}}

\definecolor{MyDarkBlue}{rgb}{0,0.08,0.5}
\definecolor{MyLightBlue}{rgb}{0,0.08,0.8}
\definecolor{MyDarkGreen}{rgb}{0.02,0.50,0.02}
\definecolor{MyDarkRed}{rgb}{0.7,0.02,0.02}
\definecolor{MyDarkOrange}{rgb}{0.40,0.2,0.02}
\definecolor{MyDarkMagenta}{rgb}{0.337,0,0.827}

\newcommand{\PMI}{\text{PMI}}

\title{Learning visual groups from co-occurrences in space and time}

\author{Phillip Isola \\
UC Berkeley\\
\texttt{phillipi@berkeley.edu}
\And
Daniel Zoran \\
MIT\\
\texttt{danielz@mit.edu}
\And
Dilip Krishnan \\
Google \\
\texttt{dilipkay@google.com}
\And
Edward H. Adelson \\
MIT\\
\texttt{adelson@mit.edu}
}

\begin{document}

\maketitle

\begin{abstract}



We propose a self-supervised framework that learns to group visual entities based on their rate of co-occurrence in space and time. To model statistical dependencies between the entities, we set up a simple binary classification problem in which the goal is to predict if two visual primitives occur in the same spatial or temporal context. We apply this framework to three domains: learning patch affinities from spatial adjacency in images, learning frame affinities from temporal adjacency in videos, and learning photo affinities from geospatial proximity in image collections. We demonstrate that in each case the learned affinities uncover meaningful semantic groupings. From patch affinities we generate object proposals that are competitive with state-of-the-art supervised methods. From frame affinities we generate movie scene segmentations that correlate well with DVD chapter structure. Finally, from geospatial affinities we learn groups that relate well to semantic place categories.

%
%
%
%
%
%


\end{abstract}


\section{Introduction}

Clown fish live next to sea anemones, lightning is always accompanied by thunder. When looking at the world around us, we constantly notice which things go with which. These associations allow us to segment and organize the world into coherent visual representations.

This paper addresses how representations like ``objects" and ``scenes" might be learned from natural visual experience. A large body of work has focused on learning these representations as a supervised problem (e.g., by regressing on image labels) and current object and scene classifiers are highly effective. However, in the absence of expert annotations, it remains unclear how we might uncover these representations in the first place. Do ``objects" fall directly out of the statistics of the environment, or are they a more subjective, human-specific construct? 

Here we probe the former hypothesis. Because the physical world is highly structured, adjacent locations are usually semantically related, whereas far apart locations are more often semantically distinct. By modeling spatial and temporal dependencies, we may therefore learn something about semantic relatedness. 

We investigate how these dependences may be learned from unlabeled sensory input. We train a deep neural network to predict whether or not two input images or patches are likely to be found next to each other in space or time. We demonstrate that the network learns dependencies that can be used to uncover meaningful visual groups. We apply the method to generate fast and accurate object proposals that are competitive with recent supervised methods, as well as to automatic movie scene segmentation, and to the grouping of semantically related photographs.

\begin{figure}[t]
 \centering
 \includegraphics[width=0.9\hsize]{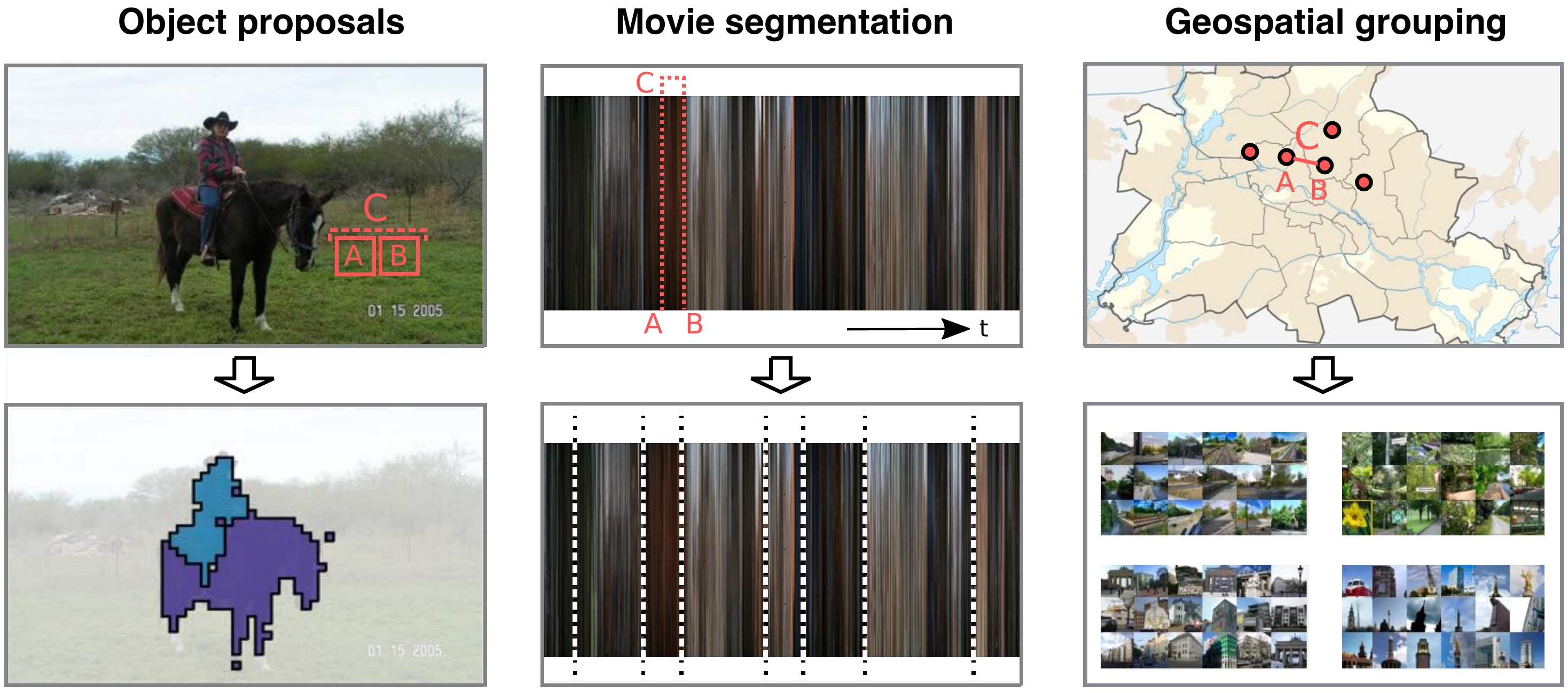}
  \vspace{-5pt}
  \caption{We model statistical dependences in the visual world by learning to predict which visual primitives -- patches, frames, or photos -- will be likely to co-occur within the same spatial or temporal context. Above, the primitives are labeled $\A$ and $\B$, and the context is labeled $\C$. By clustering primitives that predictably co-occur, we can uncover groupings such as objects (a group of patches; left), movie scenes (a group of frames; middle), and place categories (a group of photos; right).}
  \vspace{-10pt}
 \label{fig:teaser}
\end{figure}

\section{Related work}

The idea that perceptual groups reflect statistical structure in the environment has deep roots in the perception and cognition literature (\cite{barlow1985suspicious, wilkin1985perceptual, tenenbaum1983role, lowe2012perceptual, rock1983logic}). Barlow postulated that the brain is constantly on the lookout for events that co-occur much more often than chance, and uses these ``suspicious coincidences" to discover underlying structure in the world (\cite{barlow1985suspicious}). Subsequent researchers have argued that infants learn about linguistic groups from phoneme co-occurrence (\cite{saffran1996statistical}), and that humans may also pick up on visual patterns from co-occurrence statistics alone (\cite{fiser2001unsupervised, schapiro2013neural}).


Visual grouping is also a central problem in computer vision, showing up in the tasks of edge/contour detection \cite{canny1986computational, arbelaez2011contour, isola2014crisp} , (semantic) segmentation \cite{shi2000normalized, malisiewicz2007improving}, and object proposals \cite{alexe2012measuring, zitnick2014edge, krahnenbuhl2015}, among others. Many papers in this field take the approach of first modeling the affinity between visual elements, then grouping elements with high affinity (e.g., \cite{shi2000normalized}). Our work follows this approach. However, rather than using a hand-engineered grouping cue \cite{shi2000normalized,zitnick2014edge}, or learning to group with direct supervision \cite{dollar2013structured,krahnenbuhl2015}, we use a measure of spatial and temporal dependence as the affinity. 

Grouping based on co-occurrence has received some prior attention in computer vision (\cite{sivic2005discovering,faktor2012clustering,faktor2013co,isola2014crisp}). \cite{sivic2005discovering} demonstrated that object categories can be discovered and roughly localized using an unsupervised generative model. \cite{isola2014crisp} showed that statistical dependences between adjacent pixel colors, measured by pointwise mutual information ($\PMI$) can be very effective at localizing object boundaries. Both these methods require modeling generative probability distributions, which restricts their ability to scale to high-dimensional data. Our model, on the other hand, is discriminative and can be easily scaled.

A recent line of work in representation learning has taken a similar tack, training discriminative models to predict one aspect of raw sensory data from another. This work may be termed self-supervised learning and has a number of flavors. The common theme is exploiting spatial and/or temporal structure as supervisory signals. \cite{TemporalCoherence} learn a feature embedding such that features adjacent in time are similar and features far apart in time are dissimilar. \cite{srivastava2015unsupervised} predict future frames in a video, and rely on strict temporal ordering; extension to spatial or unordered data is unclear. \cite{wang2015unsupervised} use a siamese triplet loss to learn a representation that can track patches through a video. They rely on training input from a separate tracking algorithm. \cite{agrawal2015learning} as well as \cite{jayaraman2015learning} regress on egomotion signals to learn a representation. Finally, \cite{DBLP:journals/corr/DoerschGE15} learn features by predicting the relative orientation between patches within an image.

Each of these works focus on learning good generic features, which may then be applied as pre-training for a supervised model. Our current goal is rather different. Rather than learning a vector space representation of images, we search for more explicit structure, in the form of visual groups. We show that the learned groups are semantically meaningful in and of themselves. This differs from the usual approach in feature learning, where the features are not necessarily interpretable, but are instead used as an intermediate representation on top of which further models can be trained.






\begin{figure}[t]
\centering
\includegraphics[width=1.0\hsize]{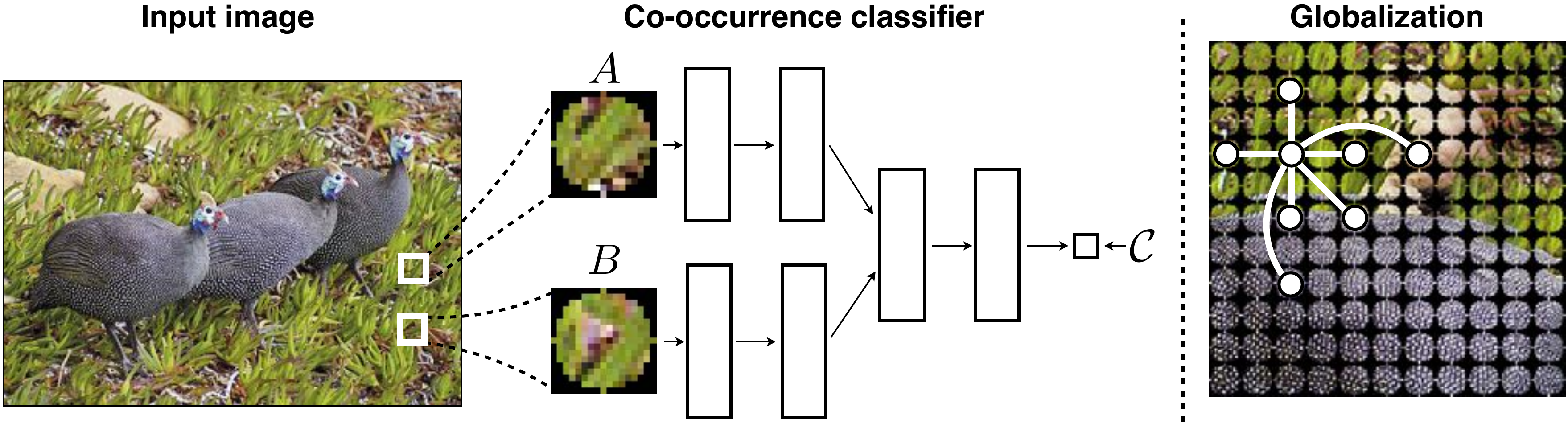}
 \caption{Overview of our approach to learning to group patches. We train a classifier to that takes two isolated patches, $\A$ and $\B$, and predicts $\C$: whether or not they were taken from nearby locations in an image. We use the output of the classifier, $P(\C=1|\A,\B)$, as an affinity measure for grouping. The rightmost panel shows our grouping strategy. We setup a graph in which nodes are image patches, and all nearby nodes are connected with an edge, weighted by the learned affinity (for clarity, only a subset of nodes and edges are shown). We then apply spectral clustering to partition this graph and thereby segment the image. The result on this image is shown in \fig{object_proposals_evaluation}.}
 \vspace{-10pt}
\label{fig:net_diagrams}
\end{figure}
\vspace{-10pt}

\section{Modeling visual affinities by predicting co-occurrence}

We would like to group visual primitives, $\A$ and $\B$, based on the probability that they belong to the same semantic entity. $\A$ and $\B$ may, for example, be two image patches, in which case we would want to group them if they belong to the same visual object. 

Given object labels, a straightforward approach to this problem would be to train a supervised classifier to predict indicator variable $\Q\in\{0,1\}$, where $\Q=1$ iff $\A$ and $\B$ lie on the same object \cite{manen2013prime}. Throughout this paper, we use $\Q$ to indicate the property that $\A$ and $\B$ share the same semantic label.

Acquiring training data for $\Q$ may require time-consuming and expensive annotation. We instead will explore an alternative strategy. Instead of training a classifier to predict $\Q$ directly, we train classifiers to predict spatial or temporal proximity, denoted by $\C\in\{0,1\}$. Because the semantics of the world change slowly over space and time, we hope that $\C$ might serve as a cheap proxy for $\Q$ (c.f. \cite{kayser2001extracting,wiskott2002slow}). The degree to which this is true is an empirical question, which we will test below. Throughout the paper, $\C=1$ iff $\A$ and $\B$ are nearby each other in space or time.

Formally, we model the affinity between visual primitives $\A$ and $\B$ as
\begin{align}
	w(\A,\B) = \frac{P(\C=1 | \A, \B) + P(\C=1 | \B, \A)}{2}.\label{eqn_1}
\end{align}
In other words, we model affinity as the probability that two primitives will co-occur within some context (with symmetry between the order of $\A$ and $\B$ enforced). We will then use this affinity metric to cluster primitives, in particular using spectral clustering (Section \ref{experiments}).

This affinity can be understood in a variety of ways. First, as described above, it can be seen as a proxy for what we are really after, $P(\Q=1 | \A, \B)$. Second, it is a measure of the spatial/temporal dependence between $\A$ and $\B$. Applying Bayes' rule we can see that $P(\C=1 | \A, \B) \propto \frac{P(\A, \B | \C=1)}{P(\A,\B)}$, which factors to $\frac{P(\A, \B | \C=1)}{P(\A)P(\B)}$ when $\A$ and $\B$ are independent. Therefore, if we sample primitives iid in space or time\footnote{In each of our applications we sample 50\% positive ($\C=1$) and 50\% negative ($\C=0$) examples. Under our logistic regression model, this results in a function monotonically related to what we would get if we had sampled iid (see appendix B.4 in \cite{king2001logistic})}, our affinity models how much more often $\A$ and $\B$ appear nearby each other than they would if they were independent (the rate of co-occurrences were they independent would simply be $P(A)P(B)$). This value is closely related to the pointwise mutual information ($\PMI$) between $\A$ and $\B$ conditioned on $\C=1$. Previous work on visual grouping \cite{isola2014crisp} and word embeddings (\cite{church1990word,levy2014linguistic}) has found PMI to be an effective measure for these tasks. 

\subsection{Predicting co-occurrences with a CNN}
To model $w(\A,\B)$ we use a Convolutional Neural Net (CNN) with a Siamese-style architecture (\fig{net_diagrams}, \cite{chopra2005learning}), which we implement in Caffe (\cite{jia2014caffe}). The network has two convolutional branches, one to process $\A$ and the other to process $\B$, with shared weights. These branches can be regarded as feature extractors. The features are then concatenated and fed to a set of fully connected layers that compare the features and try to predict $\C$. We use a logistic loss over $\C$ and train all models with stochastic gradient descent. Our objective can be expressed as
\begin{align}
	E(\mathbf{\A},\mathbf{\B},\mathbf{\C};\mathbf{\theta}) = \frac{-1}{N}\sum_1^N \mathbf{\C}_i \log (\sigma(f(\mathbf{\A}_i,\mathbf{\B}_i;\mathbf{\theta})) + (1 - \mathbf{\C}_i) \log (1 - \sigma(f(\mathbf{\A}_i,\mathbf{\B}_i;\mathbf{\theta}))
\end{align}
where $\mathbf{\theta}$ are the net parameters we optimize over (weights and biases), $N$ is the number of training examples, $\sigma$ is the logistic function, and $f$ is a neural net. For each of our experiments, $N=500,000$ training examples, 50\% of which are positive ($\C=1$) and 50\% negative ($\C=0$).

\begin{table}[t]
\label{sample-table}
\caption{Average precision scores of our method (labeled ``Co-occurrence classifier") compared to baselines at predicting $\C$ (spatial or temporal adjacency) and $\Q$ (semantic sameness) for three domains: image patches, video frames, and geospatial photos.} 
\vspace{-10pt}
\begin{center}
\begin{tabular}{lcccccc}
\multicolumn{1}{c}{}  &\multicolumn{2}{c}{\bf Patches} &\multicolumn{2}{c}{\bf Frames} &\multicolumn{2}{c}{\bf Photos} \\
\multicolumn{1}{c}{\bf Affinity measure}  &\multicolumn{1}{c}{$\C$} &\multicolumn{1}{c}{$\Q$} &\multicolumn{1}{c}{$\C$} &\multicolumn{1}{c}{$\Q$} &\multicolumn{1}{c}{$\C$} & \multicolumn{1}{c}{$\Q$}
\\ \hline 
Raw color &  0.83 & 0.73 & 0.77 & 0.58 & 0.58 &0.58 \\

Mean color &  0.87 & 0.74 & 0.82 & 0.63 & 0.56 &0.57 \\

Color histogram &  0.95 & {\bf 0.80} & 0.90 & 0.64 & 0.63 &0.62 \\

HOG &  0.67 & 0.67 & 0.77 & 0.61 & 0.63 &0.75 \\

Co-occurrence classifier &  {\bf 0.96} & {\bf 0.80} & {\bf 0.95} & {\bf 0.67} & {\bf 0.70} &{\bf 0.79} \\


\end{tabular}\label{predicting_C_and_Q_table}
\end{center}
\vspace{-10pt}
\end{table}

\begin{figure}[t]
 \centering
 \includegraphics[width=1.0\hsize]{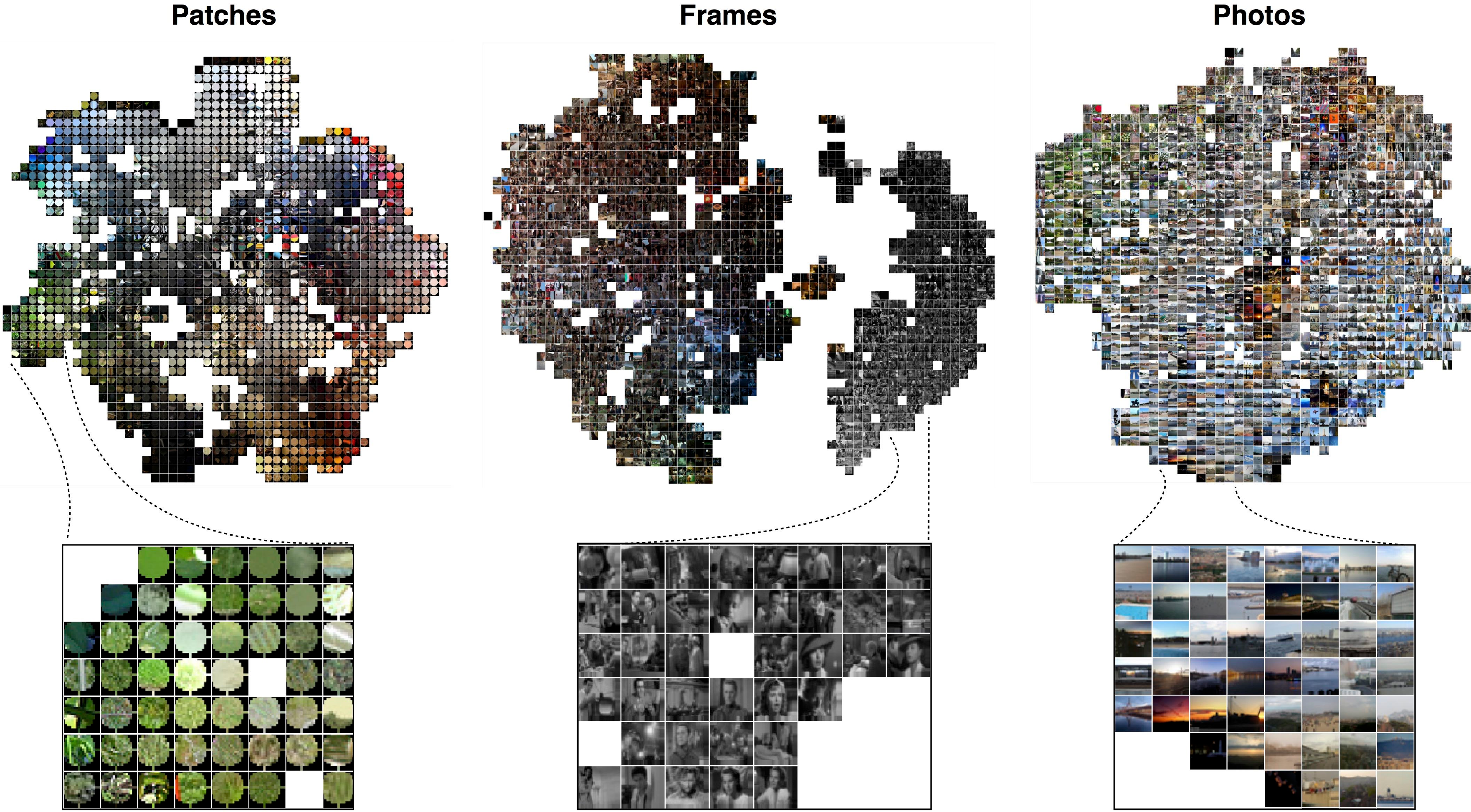}
  \caption{t-SNE visualizations of the learned affinities in each domain. We construct an affinity matrix between 3000 randomly sampled primitives to create each visualization, using $w(\A,\B)$ as the affinity measure. We then apply t-SNE on this matrix (\cite{tsne}). To avoid clutter, we visualize the embedded primitives snapped to the nearest point on a grid. The learned affinities pick up on different kinds of similarity in each domain. Patches are arranged largely according to color, while the geo-photo affinities are less dependent on color, as can be seen in the inset where day and night waterfronts map to nearby points in the t-SNE embedding.}
 \label{fig:tsnes}
\end{figure}
\vspace{-5pt}




We examine three domains: 1) learning to group patches based on their spatial adjacency in images, 2) learning to group video frames based on their temporal adjacency in movies, and 3) learning to group photos based on their geospatial proximity.

Each task corresponds to a different choice of $\A$, $\B$, $\C$, and $\Q$. In each case, we analyze performance at predicting $\C$ and at predicting $\Q$, comparing our CNN to baseline grouping cues. Each baseline corresponds to a measure of the similarity between the primitives. Similarity measures like these are commonly used in visual grouping algorithms \cite{arbelaez2011contour, faktor2012clustering}. Full results of this analysis are given in Table \ref{predicting_C_and_Q_table}. In all cases, our co-occurrence classifier matches or outperforms the baselines. 

\subsection{Patch affinities from co-occurrence in images}


We first examine whether or not predicting patch co-occurrence in images results in an effective affinity measure for object-level grouping. We set $\A$ and $\B$ to be $17\times 17$ pixel patches (with circular masks). The context function $\C$ is spatial adjacency. Positive examples ($\C=1$) are pairs of adjacent patches (with no overlap between them) and negative examples ($\C = 0$) are pairs sampled from random locations across the dataset. We sample training patches from the Pascal VOC 2012 training set. 

We train a CNN (two convolutional layers, two fully connected layers; \fig{net_diagrams}) to model $P(\C=1 | \A, \B)$. 
\fig{tsnes} shows a t-SNE visualization of the learned affinities (\cite{tsne}). As can be seen, the network learns to associate patches with different kinds of structure such as texture, local features, and color similarities.

To evaluate performance, we sample 10,000 patches from the Pascal VOC 2012 validation set, $50\%$ with $\C=1$ and $50\%$ with $\C=0$. In Table \ref{predicting_C_and_Q_table} we measure the Average Precision of using several affinity measures as a binary classifier of either $\C$ or $\Q$. In this case, we defined $\Q$ to indicate whether or not the center pixel of the two patches lies on the same labeled object instance. To test $\Q$ independently from $\C$ we create the $\Q$ test set by only sampling from patch pairs for which $\C=1$ (so the net cannot do well at predicting $\Q$ simply by doing well at predicting $\C$). Our network performs well relative to the baseline affinity metrics, although color histogram similarity does reach a similar performance on predicting $\Q$. 

Even though it was only trained to predict $\C$, our method is effective at predicting $\Q$ as well, achieving an average precision (AP) of $0.80$. This validates that spatial proximity, $\C$, is a good surrogate for ``same object", $\Q$. This raises the question, would we do any better if we directly trained on $\Q$? We tested this, training a new network on 50\% patches with $\Q=1$ and 50\% with $\Q=0$. This net achieves higher performance on predicting $\Q$ (AP = $0.85$) and lower performance at predicting $\C$ (AP = $0.92$), than our net trained to predict $\C$. Therefore, although predicting co-occurrence may be a decent proxy for predicting semantic sameness, there is still a gap in performance compared to directly training on $\Q$. Designing better context functions, $\C$, that narrow this gap is an important direction for future research.


\subsection{Frame affinities from co-occurrence in movies}
Our framework can also be applied to learning temporal associations. To test this, we set $\A$ and $\B$ to be frames, cropped and down sampled to $33\times33$ pixels, from a set of 96 movies sampled from the top 100 rated movies on IMDB\footnote{http://www.imdb.com/}. In this setting, $\C$ indicates temporal adjacency -- specifically, two frames are assigned $\C=1$ if they are at least 3 seconds from each other and not more than 10 seconds apart. $\C=0$ otherwise.

Again we train a CNN to model $P(\C=1 | \A, \B)$ (three convolutional layers, two fully connected layers). To evaluate predicting $\C$, we train on half the movies and test on the remaining half. Our method can learn to predict $\C$ quite effectively, reaching an Average Precision of $0.95$ on the test set. 

How do the learned temporal associations relate to semantic visual scenes? To test this, we compared against DVD chapter annotations, setting $\Q$ to be ``do these two frames occur in the same DVD chapter?" We sample 10,000 frame pairs, 50\% with $\Q=1$ and 50\% with $\Q=0$, while holding $\C$ constant (so that good performance at predicting $\Q$ cannot be achieved simply by doing well at predicting $\C$). Our network achieves an AP of $0.67$ on this task. Similar to above, we can then see that temporal adjacency, $\C$, is an effective surrogate for learning about semantic sameness, $\Q$.

\subsection{Photo affinities from geospatial co-occurrence}
Just as an object is a collection of associated patches, and a movie scene is a collection of associated frames, a visual \emph{place} can be viewed a collection of associated photographs. Here we set $\A$ and $\B$ to be geotagged photos, cropped and down sampled to $33\times33$ pixels, and $\C$ indicates whether or not $\A$ and $\B$ are taken within 11 meters of one another (we exclude exact duplicate locations).

Using the same CNN architecture as for the movie frame network, we again learn $P(\C=1 | \A, \B)$, but for this new setting of the variables. We train on five cities selected from the MIT City Database \cite{zhou2014cityidentity} and test predicting $\C$ on a held out set of three more cities from that dataset. We also test how well the network predicts place semantics. For this, we define $\Q$ as ``do these two photos belong to the same place category?" We test this task on the LabelMe Outdoors dataset \cite{liu2009nonparametric} for which each photo was assigned to one of eight place categories (e.g., ``coast", ``highway", ``tall building"). Our network shows promising performance on this task, reaching $0.79$ AP on predicting $\Q$. HOG similarity reaches the same performance, which corroborates past findings that HOG is effective at grouping related photos (\cite{dalal2005histograms}). 

Notice that while HOG does well on associating photographs, it does not do well at associating movie frames nor image patches. On the other hand, color histogram similarity does well on associating image patches and movie frames, but fails at grouping everyday photographs -- while patches on an object, or frames in a movie scene, may tend to all use a consistent color palette, tourist photos of the same location will have high color variance, due to seasonal and lighting variations. Different grouping rules will be effective at different tasks. Our learning based approach has the advantage that it automatically figures out the appropriate grouping cue for each new domain, and thereby achieves good performance on all our tasks.

\begin{table}[t]
\label{sample-table}
\caption{Probing the learned affinities by transforming $\B$ while leaving $\A$ unmodified. Each number reports the mean output, $w(\A,\B)$, from each network after the specified transformation has been applied. Transformations applied to one example patch are shown at the top of each column. Comparison should be made with respect to the unmodified input, given in the left-most column.} 
\vspace{-10pt}
\begin{center}
\begin{tabular}{lcccccc}
		     &  &  & Vertical  & Horizontal  & Color  & Luminance  \\
		     & No transformation & Rotated 90\degree & mirror & mirror & removed & darkened \\
		     & \protect\includegraphics[height=20pt]{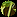} &  \protect\includegraphics[height=20pt]{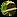}  & \protect\includegraphics[height=20pt]{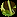}  &  \protect\includegraphics[height=20pt]{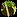}  &  \protect\includegraphics[height=20pt]{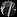}  &  \protect\includegraphics[height=20pt]{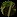}  \\
		     \hline
{\bf Patches} & 0.819  &  0.818  &  0.818  &  0.819  &  0.382  &  0.523 \\
{\bf Frames} & 0.817  &  0.794  &  0.772  &  0.813   &  0.264  &  0.608 \\
{\bf Photos} & 0.550   & 0.488   &  0.499  &  0.546   &  0.520  &  0.516 \\
\end{tabular}\label{learned_invariances}
\end{center}
\vspace{-10pt}
\end{table}

\subsection{Which cues did the networks learn to use?}
In each domain tested above, the grouping rules may be very different. Here we study them by probing the trained networks with controlled stimuli. Similar to how a psychophysicist might experiment on human perception, we show our networks specially made stimuli. For each test, we feed the networks many pairs $\{\A,\B\}$, sampled from locations such that $\C=1$. We leave $\A$ unaltered, but modify $\B$ in a controlled way. This allows us to test what kinds of transformations of $\B$ will change the network's prediction as to wether or not $\C=1$ (c.f. \cite{lenc2014understanding}). We consider the following modifications: rotation by 90\degree, mirroring vertically and horizontally, removal of color (by replacing each pixel with its mean over color channels), and a darkening transformation in which we multiply each color channel by 0.5.

Results of these tests are given in Table \ref{learned_invariances}. Each number is the mean output of the network for a given test case. The left-most column provides the mean output without any transformation applied. Interestingly, the patch network is almost entirely invariant to 90\degree rotations and mirror flips -- according to the network, sharing a common orientation does not increase the probability of two patches being nearby. On the other hand, the patch network's output depends dramatically on color similarity. The frame network behaves similarly, but shows some sensitivity to geometric transformations. On the other hand, the geo-photo network exhibits the opposite pattern of behavior: it's output is changed more when geometric transformations are applied than when color transformations are applied. According to the geo-network, two photos may have different color compositions and still be likely to be nearby.



\section{From predicting co-occurrence to forming visual groups}\label{experiments}

We apply the following general approach to learning visual groups:
\vspace{-5pt}
\begin{enumerate}
	\itemsep-1.5em
	\item Define $\A$, $\B$, and $\C$ based on the domain.\\ 
	\item Learn $P(\C | \A, \B)$ using a CNN.\\
	\item Setup a graph in which $\mathbf{\A}_{i=1}^N$ and $\mathbf{\B}_{i=1}^N$ are nodes and edge weights are given by $w(\mathbf{\A}_i,\mathbf{\B}_i)$ (Eqn. \ref{eqn_1}). Then partition the graph into visual groups using spectral clustering.\\
\end{enumerate}
\vspace{-10pt}


\subsection{Finding objects}

\begin{figure}[t]
 \centering
 \includegraphics[width=1.0\hsize]{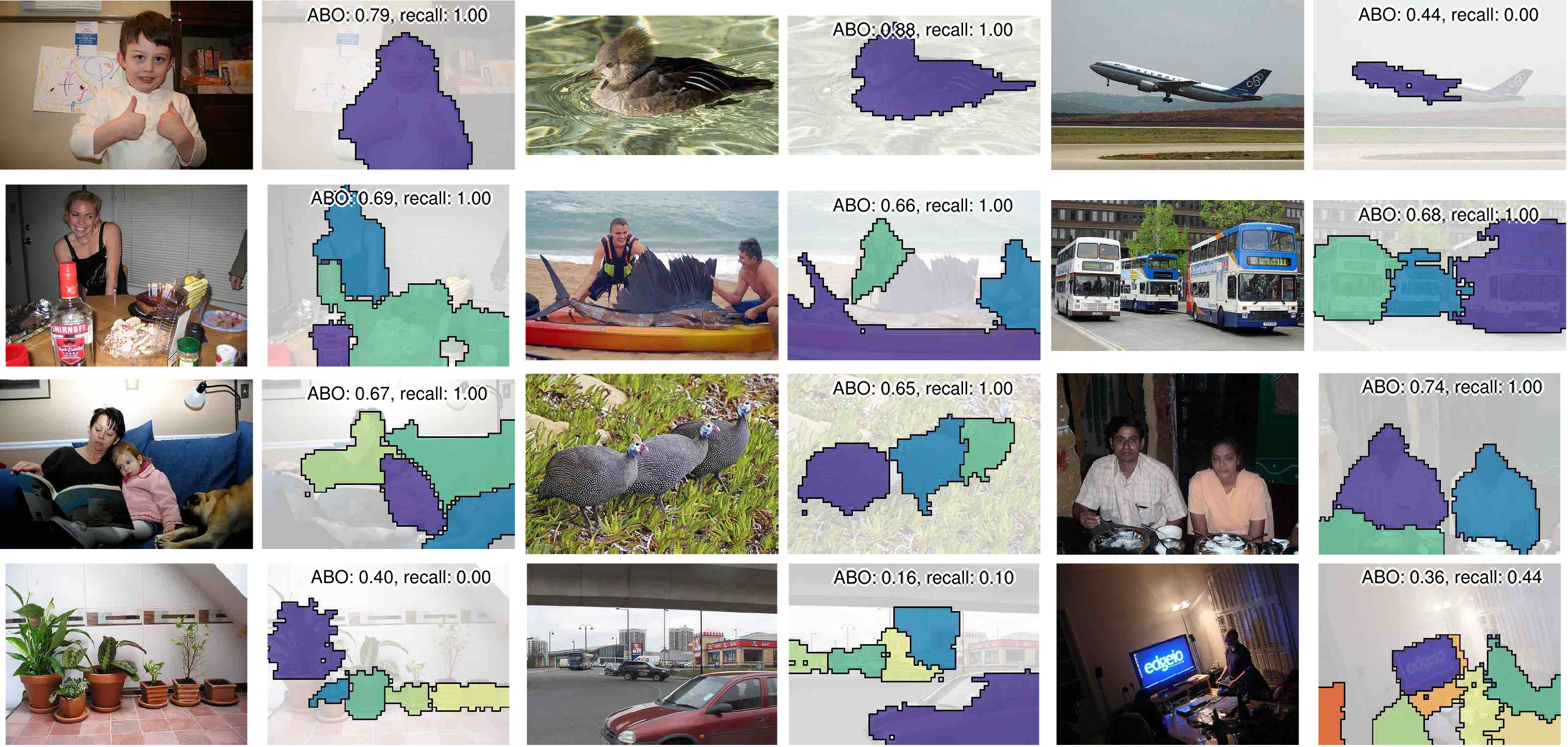}
 \vspace{-5pt}
  \caption{Example object proposals. Out of 100 proposals per image, we show those that best overlap the ground truth object masks. Average best overlap (defined in \cite{krahenbuhl2014geodesic}) and recall at a Jaccard index of 0.5 are superimposed over each result.}
  \vspace{-5pt}
 \label{fig:object_proposal_examples}
\end{figure}

\begin{figure}
 \centering
 \includegraphics[width=0.32\hsize]{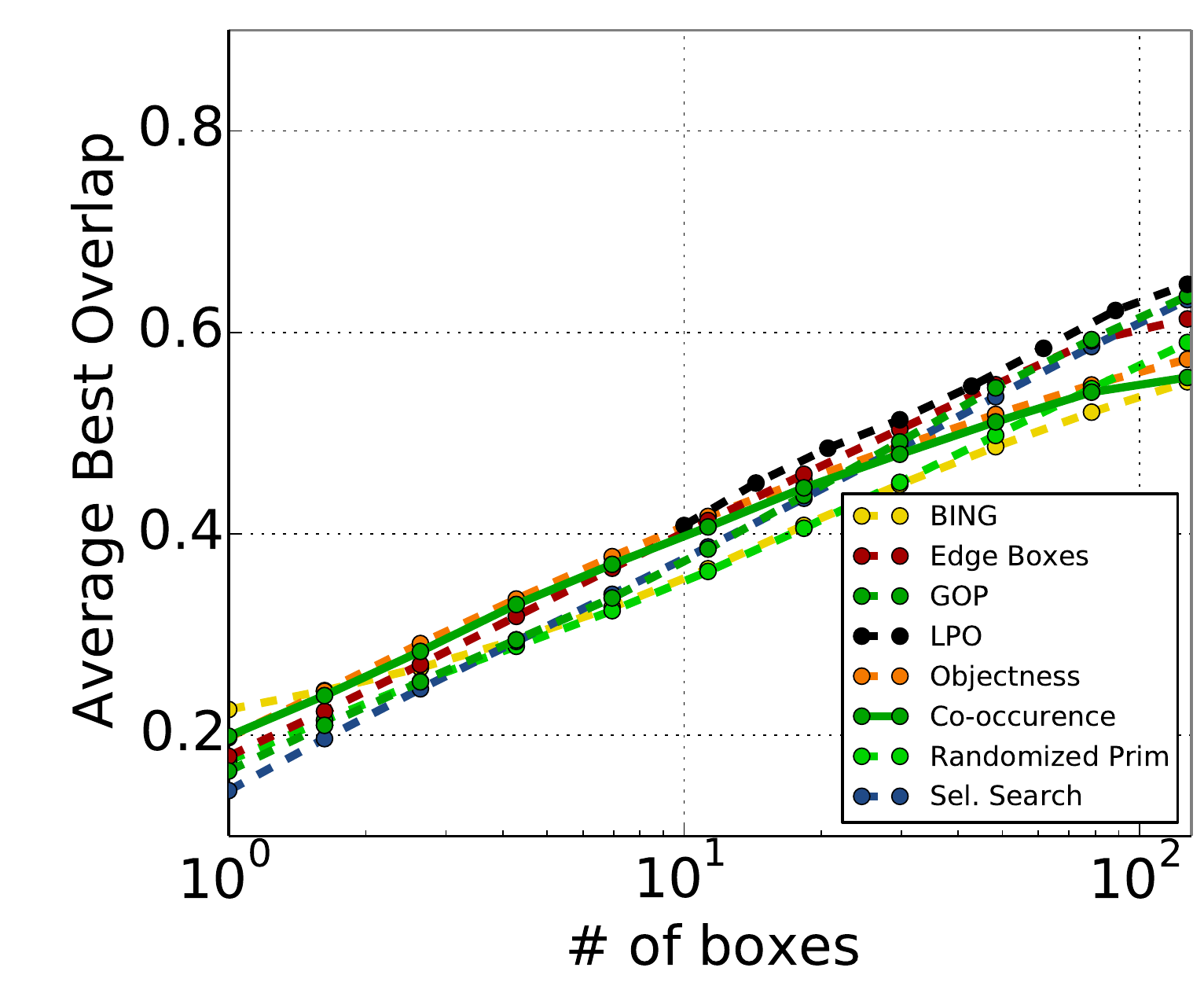}
 \includegraphics[width=0.32\hsize]{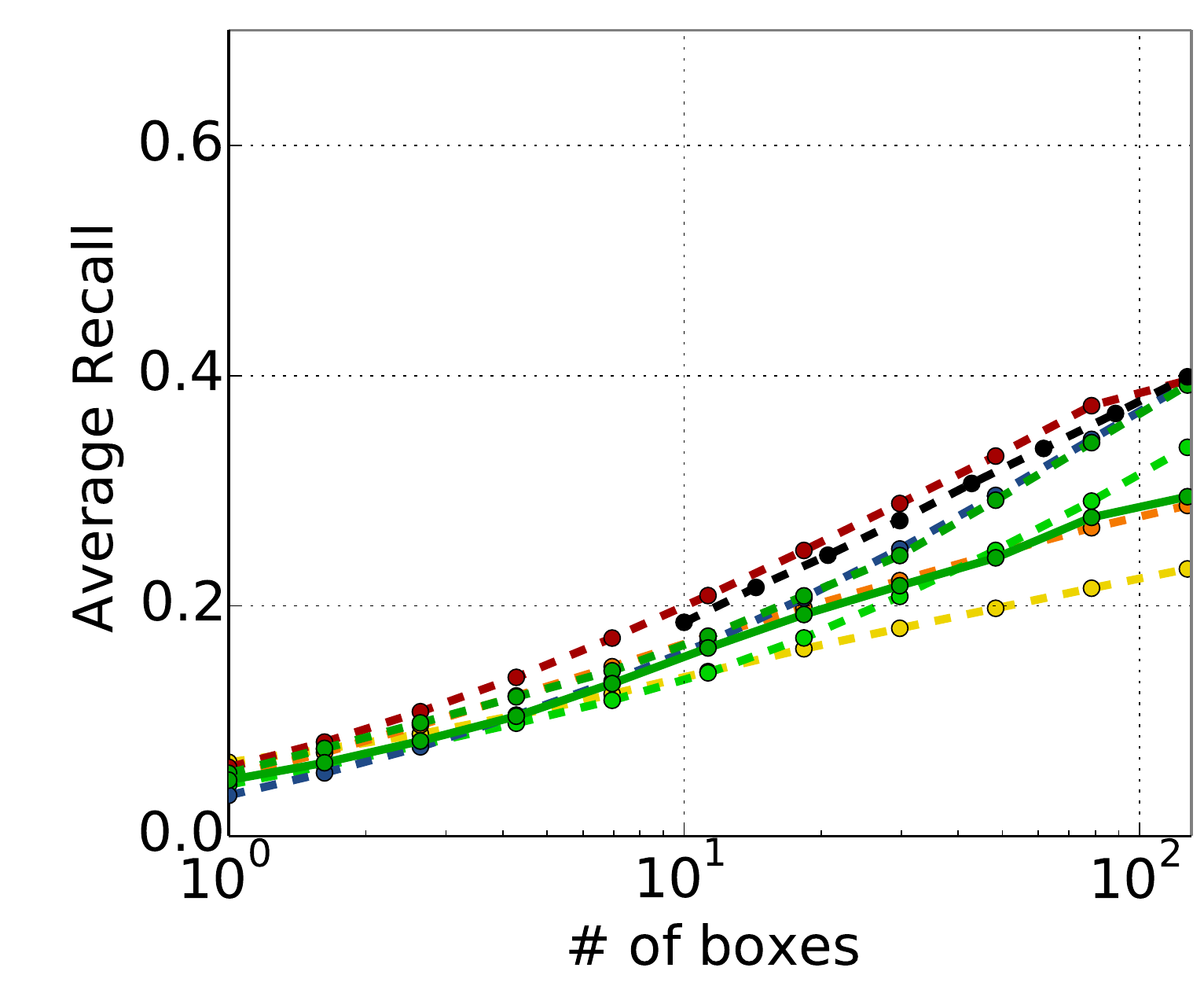}
 \includegraphics[width=0.32\hsize]{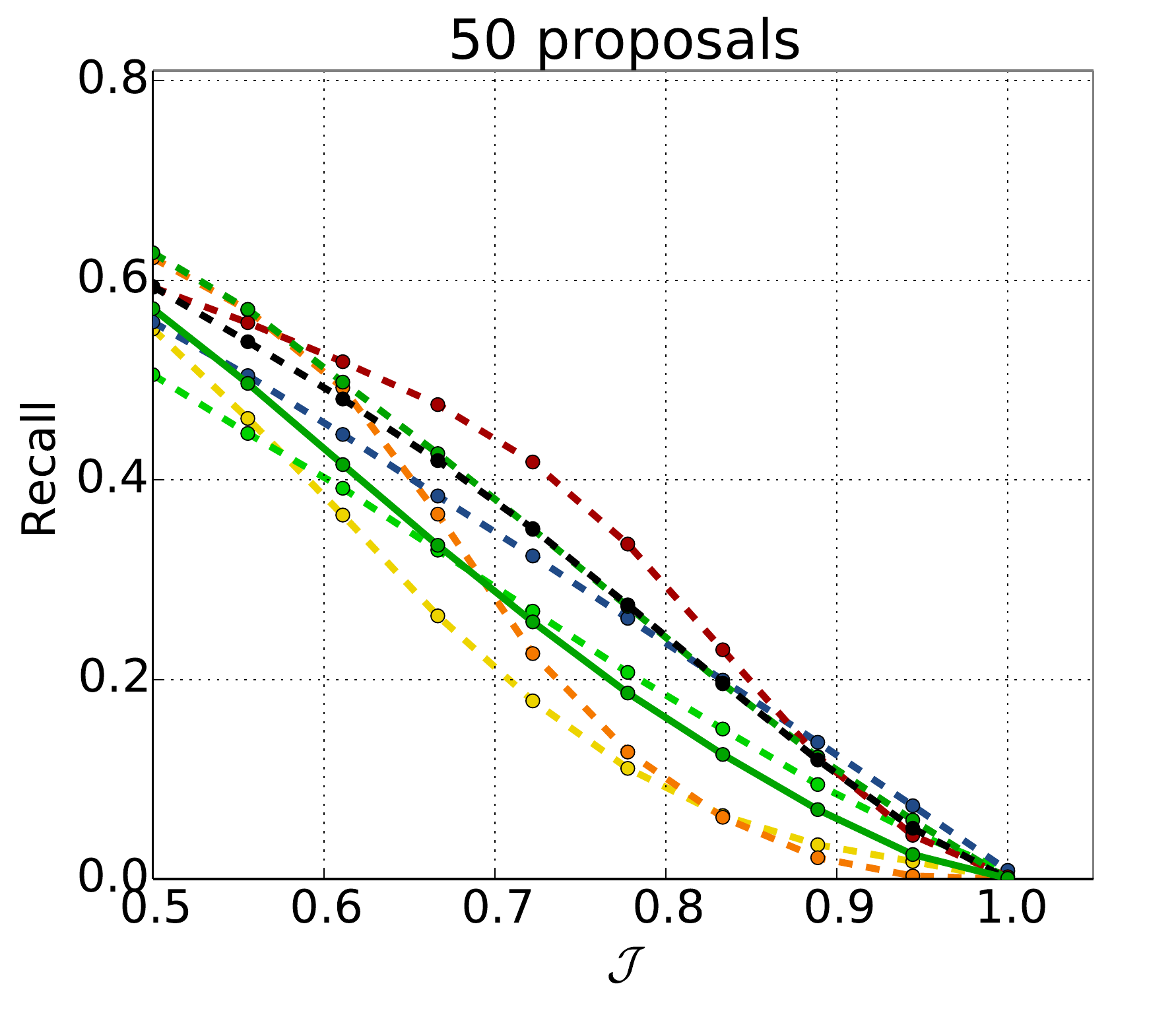}
 \vspace{-10pt}
 \caption{Object proposal results, evaluated on bounding boxes. Our unsupervised method (labeled ``Co-occurrence") is competitive with recent supervised algorithms at proposing up to around 100 objects. ABO is the average best overlap metric from (\cite{krahenbuhl2014geodesic}), $\mathcal{J}$ is Jaccard index. The papers compared to are: BING (\cite{cheng2014bing}), EdgeBoxes \cite{zitnick2014edge}, LPO (\cite{krahnenbuhl2015}), Objectness (\cite{alexe2012measuring}), GOP (\cite{krahenbuhl2014geodesic}), Randomized Prim (\cite{manen2013prime}), Sel. Search (\cite{uijlings2013selective}).}
 \vspace{-10pt}
 \label{fig:object_proposals_evaluation}
\end{figure}

As demonstrated in Section \ref{experiments}, patches that predictably co-occur usually belong to the same object. This suggests that we can localize objects in an image by grouping patches using our co-occurrence affinity. We focus on the specific problem of ``object proposals" (\cite{zitnick2014edge, krahnenbuhl2015}), where the goal is to localize all objects in an image. 

We use the patch associations $P(\C=1 | \A, \B)$ defined in Section \ref{experiments}, trained on the Pascal VOC 2012 training set.
Follow previous benchmarks, we test on the validation set. Given a test image, we sample all $17\times17$ patches at a stride of $8$ pixels. We construct a graph in which each patch is a node and nodes are connected by an edge if the spatial distance between the patch centers is at least $17$ pixels and no more than $33$ pixels. Each patch is multiplied by a circular mask so that no two patches connected by an edge see any overlapping pixels (see \fig{net_diagrams}(right)). Each edge, indexed by $i,j$, is weighted by $\mathbf{W}_{i,j} = w(\mathbf{\A}_i,\mathbf{\B}_i)^{\alpha}$, resulting in the affinity matrix $\mathbf{W}$, where we use the value $\alpha = 20$ in our experiments.

To globalize the associations, we apply spectral clustering to the matrix $\mathbf{W}$. First we create the Laplacian eigenmap $L$ for $\mathbf{W}$, using the 2nd through 16th eigenvectors with largest eigenvalues. Each eigenvector is scaled by $\lambda^{-\frac{1}{2}}$ where $\lambda$ is the corresponding eigenvalue. We then generate object proposals simply by applying k-means to the Laplacian eigenmap. To generate more than a few proposals, we run k-means multiple times with random restarts and for values of k from 5 to 16. Finally, we prune redundant proposals and sort proposals to achieve diversity throughout the ranking (by encouraging proposals to be made from different values of k before giving proposals from a random restart at the same value of k).

Qualitative results from our method are shown in \fig{object_proposal_examples}. In each case, we show the proposals that have best overlap with the ground truth object masks for 100 proposals. We quantitatively compare against other recent methods in \fig{object_proposals_evaluation}. 
Even though our method is not trained on labeled images, it reaches performance comparable to recent supervised methods at proposing up to 100 objects per image. Our implementation runs in about 4 seconds per image on a 2015 Macbook Pro. 


\subsection{Segmenting movies}

\begin{figure}[t]
 \centering
  \begin{subfigure}[b]{0.51\textwidth}
 \includegraphics[width=\textwidth]{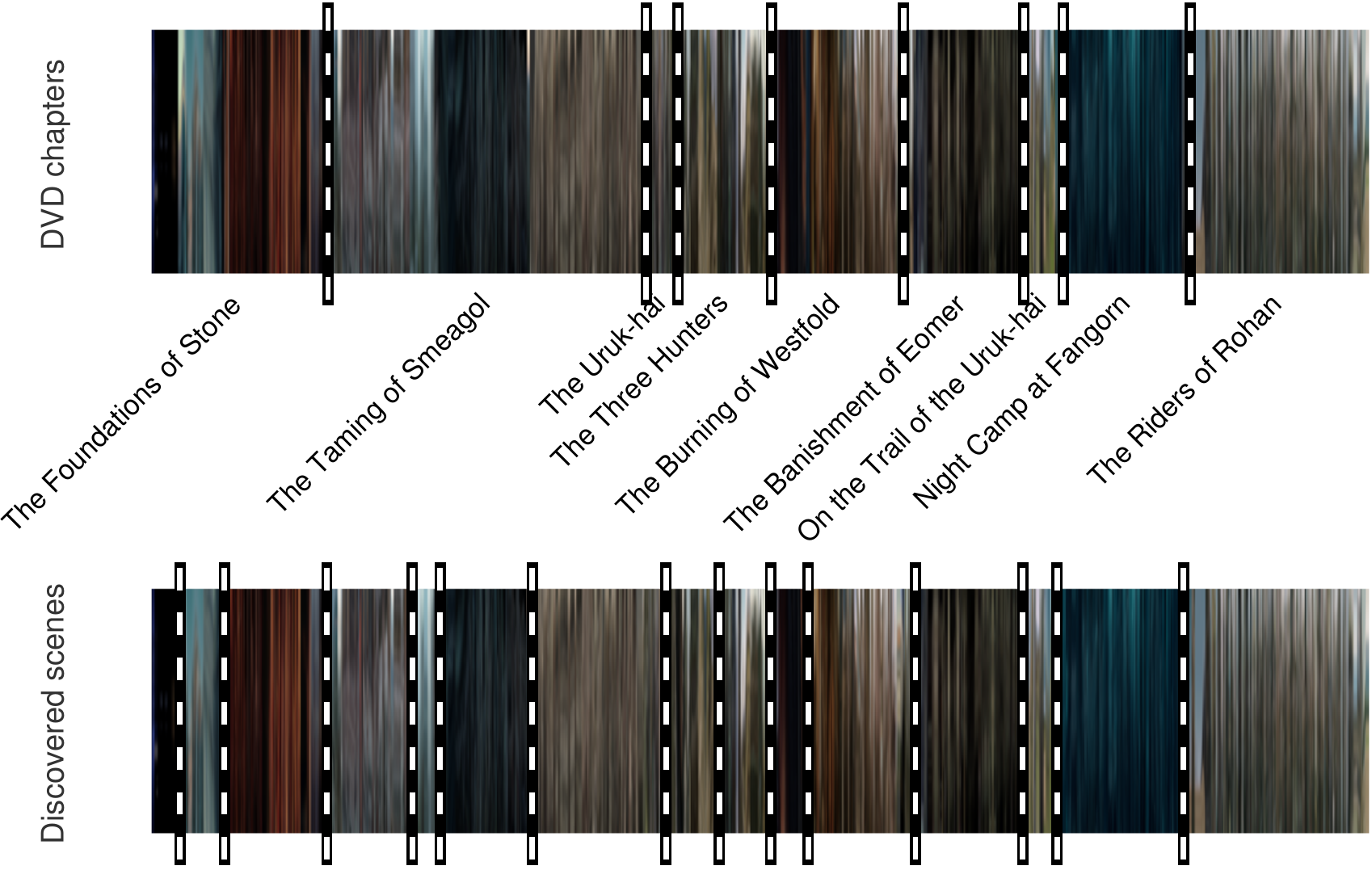}
 \end{subfigure}
 \hspace{10pt}
 \begin{subfigure}[b]{0.37\textwidth}
 \includegraphics[width=\textwidth]{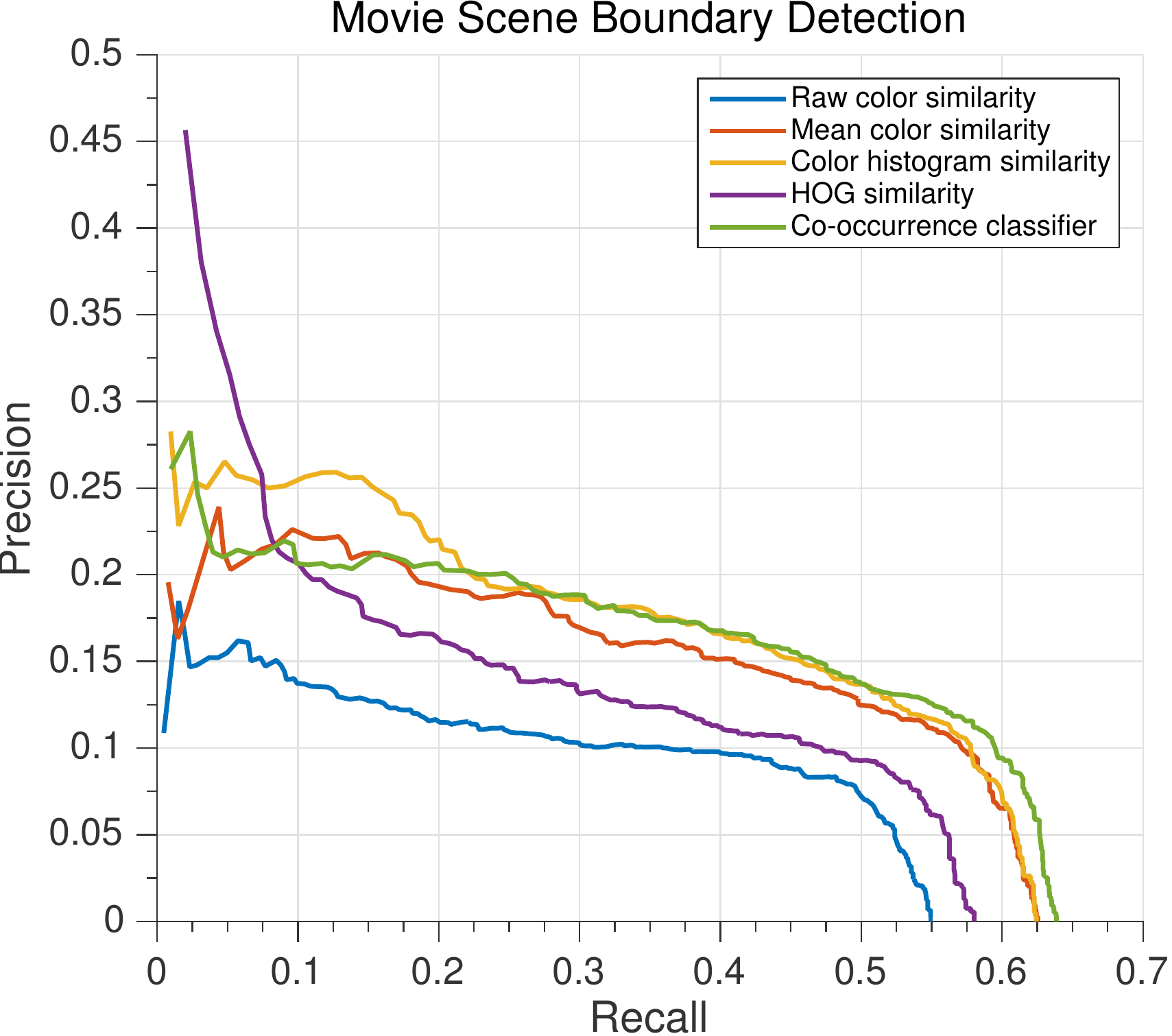}
 \end{subfigure}
  \caption{Movie scene segmentation results. On the left, we show a ``movie barcode" for \emph{The Two Towers}, in which each frame of the movie is resized into a signal column of the visualization; the top shows the DVD chapters and the bottom our recovered scene segmentation. Notice that some DVD chapters contain multiple different scenes. Our method tends to detect these sub-chapter scenes, resulting in an over-segmentation compared to the chapters. On the right, we quantify our performance on this scene segmentation task; see text for details.}
  \vspace{-10pt}
 \label{fig:scene_segmentation_results}
\end{figure}
\vspace{-10pt}

Just as objects are composed of associated patches, scenes in a movie are composed of associated frames. Here we show how our learned frame affinities can be used to break a movie into coherent scenes, a problem that has received some prior attention (\cite{chen2008movie, zhai2006video}).



To segment a movie, we build a graph in which each frame is a node and all frames within ten seconds of one another are connected by an edge. We then weight the edges using the frame-associations $P(\C=1 | \A, \B)$ (Section \ref{experiments}), and partition the graph using spectral clustering.

\begin{figure}
 \centering
 \begin{subfigure}[t!]{0.56\textwidth}
 \includegraphics[width=\textwidth]{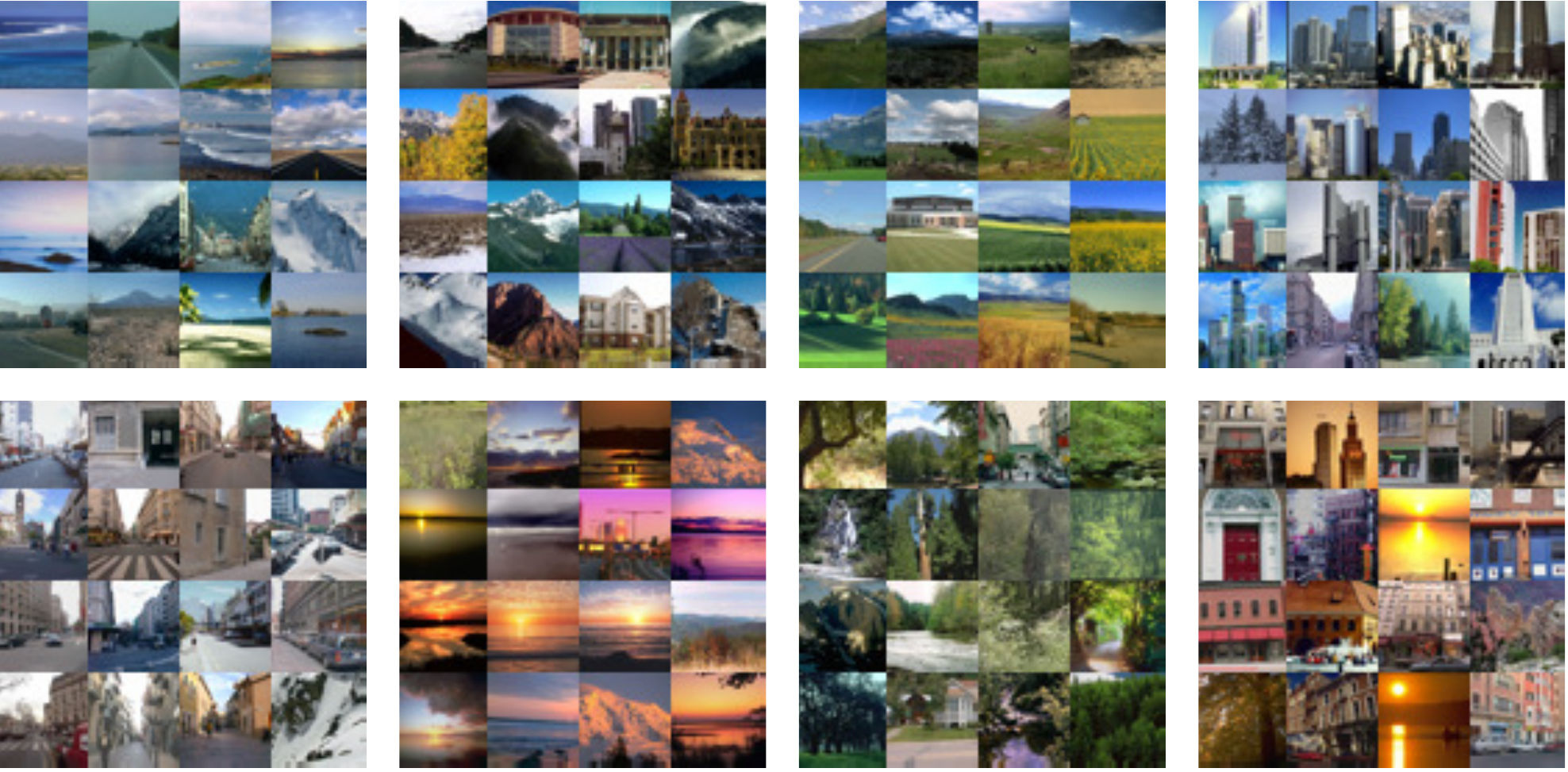}
 \end{subfigure}
 \hspace{10pt}
 \begin{subfigure}[t!]{0.37\textwidth}
 \includegraphics[width=\textwidth]{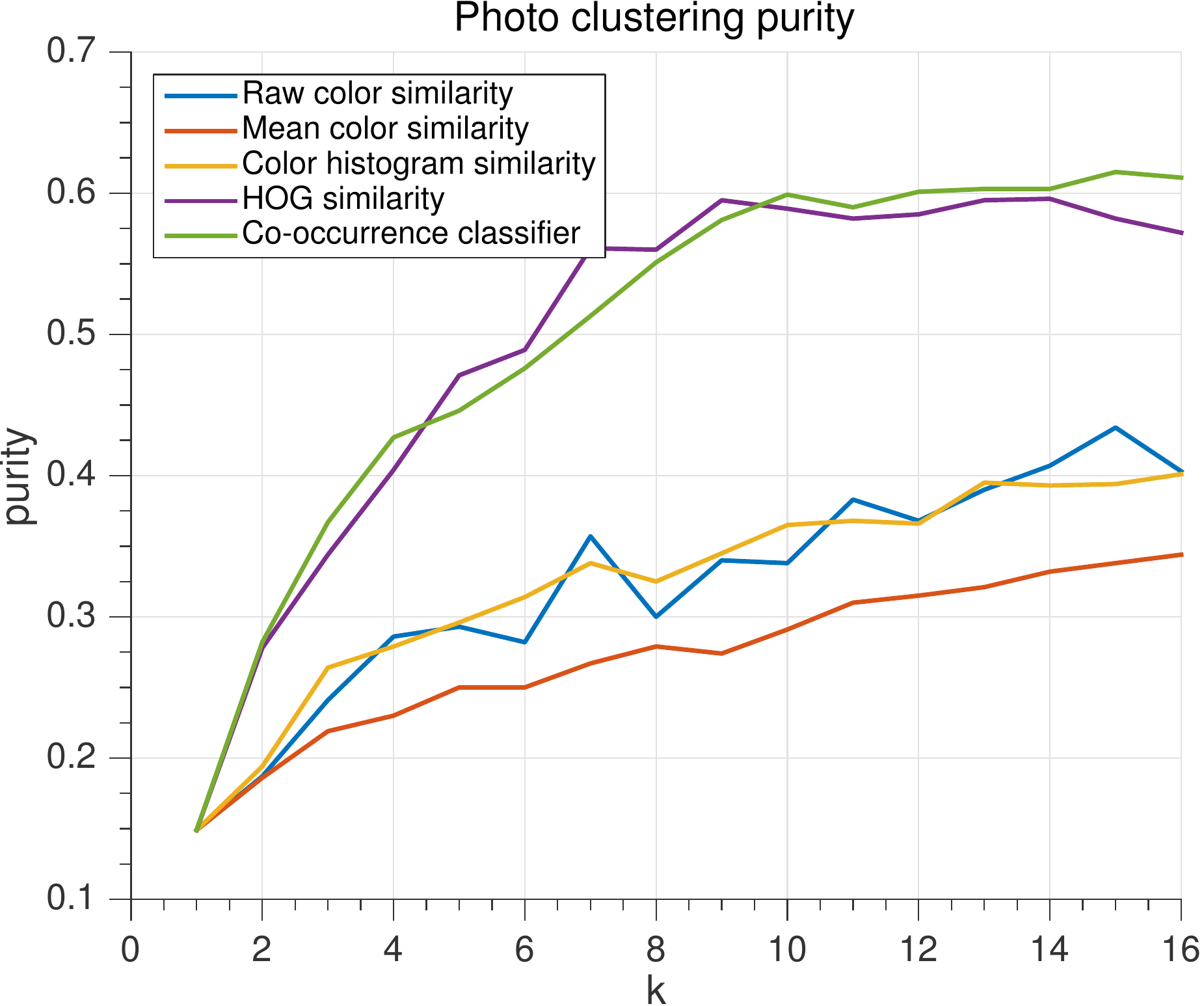}
 \end{subfigure}
 \vspace{-5pt}
  \caption{Left: Clustering the LabelMe Outdoor dataset (\cite{liu2009nonparametric}) into 8 groups using our learned affinities. Random sample images are shown from each group. Right: Photo cluster purity versus number of clusters $k$. Note that we trained our model on an independent dataset, the MIT City dataset (\cite{zhou2014cityidentity}).}
  \vspace{-10pt}
 \label{fig:geo_clustering}
\end{figure}

To evaluate, we use DVD chapter annotations as ground truth. Following a standard evaluation procedure in image boundary detection (\cite{arbelaez2011contour}), we measure performance on the retrieval task of finding all ground truth boundaries. In \fig{scene_segmentation_results}(right), we compare against the baseline affinity metrics from Section \ref{experiments}. In each case, we apply the same spectral clustering pipeline as for our method, except with edge weights given by the baseline metrics instead of by our net. It is important to properly scale each similarity metric or else spectral clustering will fail. To provide fair comparison, we sweep over a range of scale factors $\alpha$, setting edge weights as $\exp(\frac{w(\A,\B)}{\alpha^2})$, where $w$ is the affinity measure. In \fig{scene_segmentation_results}(right) we show results selecting the optimal $\alpha$ for each method.

Our approach finds more boundaries with higher precision than these baselines, except for color histogram similarity, which reaches a similar performance. \fig{scene_segmentation_results} (left) shows an example segmentation of a section of \emph{The Two Towers}. The movie is displayed as a ``movie barcode"\footnote{http://moviebarcode.tumblr.com/} in which each frame is squished into a single column and time advances to the right. On top are the DVD chapter annotations, and on the bottom are our inferred boundaries. 

\subsection{Discovering place categories}
Taking the geospatial-associations model from Section \ref{experiments}, we cluster photos into coherent types of places. Here we create a fully connected graph between all photos in a given collection, weight the edges with $P(\C=1 | \A,\B)$ and then apply spectral clustering to partition the collection. We test the purity of the clusters on LabelMe Outdoors dataset (\cite{liu2009nonparametric}). Clustering purity versus number of clusters $k$ is given in \fig{geo_clustering} (right), showing that our method is effective at discovering semantic place categories. As in our movie segmentation experiments, we select the optimal $\alpha$ to scale the affinity of our method as well each baseline. \fig{geo_clustering} (left) shows random sample images from each cluster after clustering into 8 categories. This clustering has $59\%$ purity. 




\section{Conclusion}
We have presented a simple and general approach to learning visual groupings, which requires no pre-defined labels. Instead our framework uses co-occurrence in space or time as a supervisory signal. By doing so, we learn different clustering mechanisms for a variety of tasks. Our approach achieves competitive results on object proposal generation, even when compared to methods trained on labeled data. Additionally, we demonstrated that the same method can be used to segment movies into scenes and to uncover semantic place categories. The principles underlying the framework are quite general and may be applicable to data in other domains, when there are natural co-occurrence signals and groupings. 

\subsubsection*{Acknowledgments}
We thank William T. Freeman, Joshua B. Tenenbaum, and Alexei A. Efros for helpful feedback and discussions. Thanks to Andrew Owens for helping collect the movie dataset. This work is supported by NSF award 1161731, Understanding Translucency, and by Shell Research.

{\small
\bibliography{learning_visual_associations2}
\bibliographystyle{iclr2016_conference}
}

\end{document}